\documentclass{article}

\PassOptionsToPackage{numbers}{natbib}
\usepackage[preprint]{neurips_2026}

\usepackage[utf8]{inputenc} 
\usepackage[T1]{fontenc}    
\usepackage{hyperref}       
\usepackage{url}            
\usepackage{booktabs}       
\usepackage{amsfonts}       
\usepackage{nicefrac}       
\usepackage{microtype}      
\usepackage{xcolor}         

\usepackage{booktabs}
\usepackage{multirow}
\usepackage{colortbl}
\usepackage{xcolor}
\usepackage{makecell}
\usepackage{graphicx}
\definecolor{lightgreenbg}{rgb}{0.85,0.98,0.85}
\definecolor{diffgray}{rgb}{0.45,0.45,0.45}
\newcommand{\diff}[1]{%
  \colorbox{lightgreenbg}{\textcolor{diffgray}{\scriptsize $\uparrow$ #1}}%
}
\usepackage{makecell}
\usepackage{tcolorbox}
\usepackage{amsmath}
\newcommand{\ours}{{\fontfamily{ppl}\selectfont CoE-Agent}}

\title{\ours: Co-Evolving Patient-Doctor Agents via\\ Interactive Policy Graph Optimization \\for Clinical Decision Making}

\author{%
  Guolin Huang \\
  College of Computer Science \\
  and Software Engineering, \\
  Shenzhen University \\
  Shenzhen, China \\
  \texttt{a1760050788@163.com}
  \And
  Wenting Chen\footnotemark[1] \\
  Department of Radiation Oncology, \\
  Stanford University \\
  Stanford, California, USA \\
  \texttt{wentingchen168@gmail.com}
  \And
  Linlin Shen\thanks{Corresponding authors.} \\
  Intelligent Computing Center, \\
  School of AI, \\
  Shenzhen University \\
  Shenzhen, China \\
  \texttt{llshen@szu.edu.cn}
}

\begin{document}

\maketitle

\begin{abstract}
  Clinical Decision Making (CDM) requires integrating heterogeneous information across sequential diagnostic stages while interacting with patients whose behaviors are often non-stationary and unreliable. However, existing medical agents are typically trained against cooperative patient simulators and tackle each clinical task in isolation, making them fragile to dynamic patient behaviors and unable to capture inter-task dependencies essential for globally coherent decision-making. 
  To address these issues, we propose~\ours, a co-evolving patient–doctor agentic framework that learns globally coherent CDM strategies via interactive policy graph optimization. First, Dynamic Patient–Doctor Co-Evolutionary Learning pairs a Dynamic Patient Agent, which maintains an evolving cognition state to simulate non-stationary behaviors, with an Interactive Policy Graph-Based Doctor Agent that generates DAG-structured policy graphs unifying interactions, tool invocations, and reasoning steps. The doctor agent is then optimized via Interactive Policy Graph Optimization (IPGO), which jointly enforces structural validity and task-aligned correctness through reinforcement learning. Second, Cross-Task Policy Graph Consolidation and Unified Model Learning prunes low-value nodes, aggregates high-value nodes across tasks while filtering redundancy, and distills a unified policy planner with task-specific decision models for globally coherent decision-making. Experiments on MedChain and ClinicalBench show that \ours~surpasses state-of-the-art baselines by 11.13\% and 23.63\% in average score, while achieving up to 19.5$\times$ faster inference. The source code is available at \url{https://github.com/hotdogisme/CoE-Agent}.

\end{abstract}

\section{Introduction}

Clinical Decision Making (CDM) stands as one of medicine's most complex challenges at the intersection of AI and healthcare. It requires integrating diverse data sources, continuously assessing evolving clinical scenarios, and making evidence-based diagnostic and treatment judgments. Recent advances in large language models (LLMs) have spurred the development of medical agents to assist this process. On one hand, multi-agent systems such as MedChain-Agent~\cite{liu2025medchainbridginggapllm} and ClinicalAgent~\cite{yan2024clinicallab} explicitly model sequential diagnostic procedures and complex clinical tasks through collaborative reasoning among specialized roles. On the other hand, self-evolving paradigms have emerged, with frameworks like AgentHospital~\cite{li2025agenthospitalsimulacrumhospital} and MedAgentSim\cite{almansoori2025selfevolvingmultiagentsimulationsrealistic} introducing simulated clinical environments where agents autonomously refine their diagnostic strategies and medical expertise through repeated interactions. Despite their promise in streamlining clinical workflows, these medical agents still face some challenges.
%


Although existing medical agents perform well only when interacting with cooperative, well-behaved patients, they struggle when \textit{patient behavior becomes unstable or unpredictable}. In clinical practice, patients may misunderstand questions, misreport symptoms, provide inconsistent information, or express strong preferences and emotions~\cite{schuermeyer2017patients}. Such misreporting is far from rare; in fact, it occurs in approximately 15–20\% of real-world cases~\cite{fleischer2015inattentive,merckelbach2019patients}, often introducing non-negligible interference into clinical conversations~\cite{vanderford2001communication}. Recognizing the significance of this issue, several recent studies have begun to address it. For instance, Li~\textit{et al.} \cite{li2026idealizedpatientsevaluatingllms} introduce CPB-Bench, a medical benchmark designed to evaluate LLM performance under challenging patient behaviors, revealing that current medical LLMs consistently fall into recurring failure patterns, particularly when confronted with contradictory patient information and abnormal clinical values. To mitigate this, systems like PaMis have been proposed to identify misreports via clarifying questions, while frameworks such as EvoPatient~\cite{du-etal-2025-llms} focus on improving the factual consistency of simulated patient responses. However, these efforts are largely directed at improving the patient simulator rather than strengthening the doctor model's ability to adapt to dynamic patient, which leaves the doctor model insufficiently robust. Therefore, it is essential for current medical agents to learn more robust decision-making strategies that can accommodate \textit{\textbf{dynamic patient behaviors}}.

Another challenge lies in the fact that current medical agents tackle each task in the CDM process independently, leading to \textbf{\textit{suboptimal solutions for the entire clinical workflow}}. 
 For instance, MedChain-Agent~\cite{liu2025medchainbridginggapllm} adds a feedback mechanism at each stage to reduce error propagation, while ClinicalAgent~\cite{yan2024clinicallab} decomposes diagnosis into six stages and assigns relevant departments to optimize each locally. In real-world practice, CDM is an inherently sequential and interdependent process, in which the outcome of each task is contingent upon the preceding ones, and the quality of a downstream task can in turn reveal deficiencies in upstream decisions. When unexpected findings emerge in later stages (e.g., inconclusive laboratory results or contradictory imaging evidence), clinicians often {roll back} to revisit earlier steps, such as re-collecting the patient's history or ordering additional examinations. As a result, the responses across different tasks are deeply intertwined and should be {jointly} optimized rather than treated in isolation. However, existing medical agents for CDM scenarios mainly resolve each task in a forward-only manner, ignoring the bidirectional dependencies among tasks and thereby yielding suboptimal end-to-end solutions. Thus, it is essential to explicitly \textit{\textbf{model the inter-task relationships}} and design a \textit{\textbf{globally coherent decision-making strategy}} that supports dynamic revision and information flow across the entire diagnostic pipeline.

To address these challenges, we propose~\textbf{\ours}, a co-evolving patient-doctor agentic framework that learns globally coherent clinical decision-making strategies through interactive policy graph optimization, comprising two complementary stages (Fig.~\ref{fig:main}). In the first stage, \textit{Dynamic Patient–Doctor Co-Evolutionary Learning} lets the doctor agent optimize against an evolving patient agent to handle \textbf{\textit{dynamic patient behaviors}}. Specifically, we construct a \textit{Dynamic Patient Agent} that maintains a continuously evolving cognition state, which is initialized from the case summary, updated through doctor-patient interactions, and transferred across tasks, thereby faithfully simulating the non-stationary and potentially unreliable nature of real patients. In parallel, an \textit{Interactive Policy Graph-Based Doctor Agent}, consisting of a policy planner, a critic model, and a decision model, generates a directed acyclic graph (DAG)-structured policy graph that unifies patient interactions, tool invocations, and reasoning steps for each clinical task. To optimize the policy planner against the dynamic patient agent, we further design \textit{Interactive Policy Graph Optimization (IPGO)}, which formulates the policy graph generation as a reinforcement learning problem. IPGO jointly leverages a Policy Graph-Grounded Reward to regularize the structural validity of the policy graph (e.g., node scale and tool-node correctness) and a Task-Aligned Reward to align the executed graph with the final task outcome. In the second stage, \textit{Cross-Task Policy Graph Consolidation and Unified Model Learning} is designed to \textbf{\textit{model inter-task relationships}} and yield a \textbf{\textit{globally coherent strategy}} across the entire clinical workflow. Concretely, \textit{Reward-Based Policy Graph Pruning} first removes low-contribution nodes from each task-specific policy graph based on their marginal reward contribution; \textit{Unified Policy Graph Construction} then aggregates high-value nodes across tasks, filters semantically redundant ones via inter-node similarity, and assembles a unified policy graph that captures cross-task dependencies; finally, \textit{Unified Policy Planning and Decision Learning} performs supervised fine-tuning to obtain a unified policy planner together with task-specific decision models, enabling consistent and globally optimized decision-making across the entire CDM pipeline. Extensive experiments on two CDM benchmarks, i.e. MedChain and ClinicalBench, show that CoE-Agent significantly outperforms state-of-the-art models, validating its robustness to dynamic patient interactions and its effectiveness in globally coherent CDM.
Our contributions are summarized as follows:
\begin{itemize}
    \item We propose \ours, a co-evolving patient–doctor agentic framework that learns globally coherent clinical decision-making strategies via interactive policy graph optimization.
    
    \item For {dynamic patient behaviors}, we design a {dynamic patient–doctor co-evolutionary learning} stage, where doctor agent is iteratively improved against patient agent via reinforcement learning for robust decision-making. To consider {inter-task dependencies}, we further introduce a {cross-task policy graph consolidation and unified model learning} stage to aggregate task-specific policy graphs into a unified one for globally coherent decision-making across the entire CDM pipeline.
    
    \item Extensive experiments on two CDM benchmarks shows the superiority of \ours over the state-of-the-art medical agents.
\end{itemize}

\section{Related Works}

\subsection{Medical Agents in Clinical Decision Making}
Existing LLM-based medical agents for clinical decision-making (CDM) fall into two paradigms: \textit{multi-agent collaboration}, where specialized roles interact to solve medical tasks (e.g. ClinicalAgent~\cite{yan2024clinicallab}, MedChain-Agent~\cite{liu2025medchainbridginggapllm}), and \textit{self-evolving simulation}, where doctor agents improve through repeated interactions in simulated hospitals (e.g., Agent Hospital~\cite{li2025agenthospitalsimulacrumhospital}, MedAgentSim~\cite{almansoori2025selfevolvingmultiagentsimulationsrealistic}). However, these methods handle clinical stages in isolation with forward-only information flow and assume cooperative patients, making them brittle to the misreporting and contradictions typical in real consultations~\cite{fleischer2015inattentive,merckelbach2019patients}. In contrast, \ours~co-evolves a dynamic patient agent with an interactive policy-graph-based doctor agent via reinforcement learning, unifying task-specific policy graphs into a globally coherent CDM strategy that is robust to non-stationary patient behaviors and aware of cross-task dependencies.

\subsection{Self-Evolving Agents}
Self-evolving agents have been widely investigated and can be categorized based on their evolution targets into four groups, including models, context, tools, and architectures. First, agents evolve models by updating internal parameters; for example, SCA~\cite{zhouself} fine-tunes weights via self-generated tasks, while TextGrad~\cite{yuksekgonul2024textgrad} uses textual feedback for policy improvement. Second, agents adapt their context through memory and prompt optimization. Systems like Mem0~\cite{chhikara2025mem0buildingproductionreadyai} and Expel~\cite{zhao2024expel} dynamically update knowledge and rules. Third, agents expand their tools by synthesizing new capabilities, as demonstrated by Voyager's~\cite{wangvoyager} autonomous skill library and SkillWeaver's~\cite{zheng2025skillweaver} API generation. Finally, agents optimize their agentic architectures and workflows. This ranges from single-agent code self-modification, like the Darwin Gödel Machine~\cite{Zhang2025DarwinGM}, to multi-agent workflow optimization frameworks such as AFlow~\cite{zhang2025aflowautomatingagenticworkflow}. Despite these advancements, these self-evolving agents primarily optimize isolated capabilities in static or cooperative environments, struggling with dynamic patient behaviors and bidirectional inter-task dependencies. To bridge this gap, \ours~extends the self-evolving paradigm into a \textit{co-evolutionary} setting. By optimizing an interactive policy graph via reinforcement learning against a dynamic patient agent, \ours~adapts to unpredictable behaviors and consolidates task-specific graphs into a globally coherent clinical strategy.

\begin{figure*}[t]
	\centering
	\includegraphics[width=\textwidth]{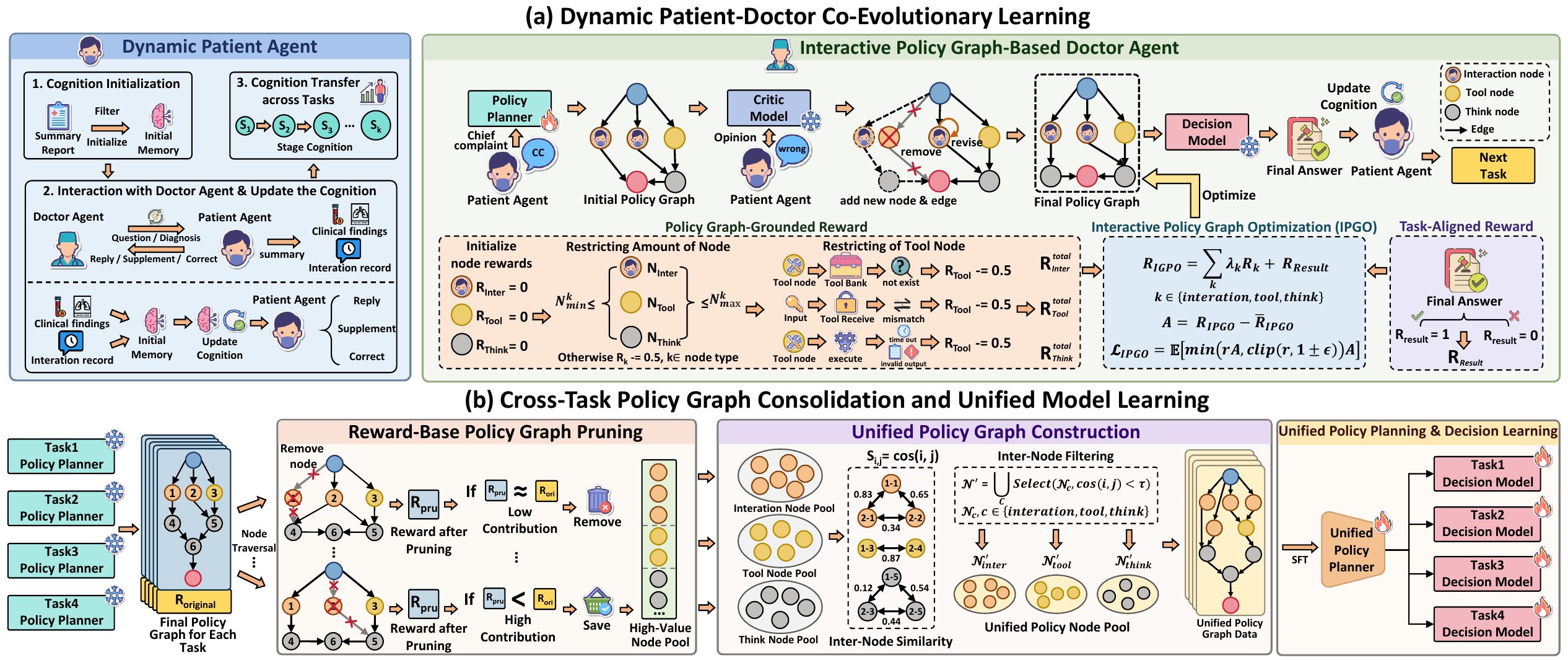}	
	\caption{\textbf{Overview of \ours}, consisting of (a) \textbf{\textit{dynamic patient-doctor co-evolutionary learning}} to optimize task-specific policy graphs through patient-doctor interaction, and (b) \textbf{\textit{cross-task policy graph consolidation and unified model learning}} to build a unified policy planner for multi-task clinical decision-making}
	\label{fig:main}
\end{figure*}

\section{Method}

\noindent\textbf{Overview.}
In Fig.~\ref{fig:main}, \ours~is a co-evolving patient–doctor agentic framework that aims to acquire robust and globally coherent CDM strategies through interactive policy graph optimization, including two complementary stages. In the first stage, \textit{Dynamic Patient-Doctor Co-Evolutionary Learning} (Sec.~\ref{sec:stage1}) constructs a \textit{Dynamic Patient Agent} that simulates the dynamic behaviors of real patients, and an \textit{Interactive Policy Graph-Based Doctor Agent} that produces a directed acyclic graph (DAG)-structured policy graph for each clinical task. The doctor agent is co-evolved with the patient agent through \textit{Interactive Policy Graph Optimization (IPGO)}, which jointly leverages a Policy Graph-Grounded Reward and a Task-Aligned Reward to ensure both structural validity and task-aligned correctness of the policy graph. In the second stage, \textit{Cross-Task Policy Graph Consolidation and Unified Model Learning} (Sec.~\ref{sec:stage2}) prunes low-value nodes, aggregates high-value nodes across tasks into a unified policy graph, and finally trains a unified policy planner with task-specific decision models via SFT, yielding a globally coherent strategy spanning the entire CDM pipeline.

\subsection{Dynamic Patient-Doctor Co-Evolutionary Learning (CoEL)}
\label{sec:stage1}
To mitigate the challenge of \textit{dynamic patient behaviors}, CoEL models the patient as a {continuously evolving} environment and iteratively optimizes the doctor agent against it. It consists of a \textit{Dynamic Patient Agent} that maintains an updatable cognition state to simulate non-stationary patient behaviors, and an \textit{Interactive Policy Graph-Based Doctor Agent} that generates DAG-structured policy graphs and is co-evolved with the patient via Interactive Policy Graph Optimization (IPGO).

\subsubsection{Dynamic Patient Agent (DPA)}

The dynamic patient agent is built around an updatable {cognition state} $c_t$, which represents the patient's current understanding of the symptoms, prior examinations, and overall condition at time step $t$. The cognition state is initialized from the basic case information available at the first visit, continuously updated through doctor–patient interactions, and transferred across clinical tasks.

\noindent\textbf{Cognition Initialization.}
To prevent information leakage and reflect the limited data available at first visit, \(c_t\) is initialized using only the chief complaint and basic medical history, excluding any unobservable future information:
$c_0 = \mathcal{I}(x_{\text{case}})$,
where \(x_{\text{case}}\) is the full case record and \(\mathcal{I}(\cdot)\) filters for the subset observable at admission.

\noindent\textbf{Interaction with Doctor Agent and Cognition Update.}
During doctor–patient interaction, the patient's cognition is continuously reshaped by the doctor's actions, including questions, examination orders, and diagnostic conclusions. Let $a_t$ denote the doctor's action and $o_t$ the patient's response at step $t$. The cognition is updated as
$c_{t+1} = \mathcal{U}(c_t, a_t, o_t) + \epsilon_t$,
where $\mathcal{U}(\cdot)$ is the cognition update function and $\epsilon_t$ captures drift from misdiagnoses or unreasonable exam requests. Thus, flawed decisions bias subsequent feedback, while sound ones drive $c_t$ toward a faithful state. Conditioned on the updated cognition, the patient agent chooses to \textit{reply}, \textit{supplement}, or \textit{correct} the doctor, yielding realistic non-stationary behavior.


\noindent\textbf{Cognition Transfer Across Tasks.}
CDM consists of multiple sequential tasks $\{T_1, T_2, \ldots, T_K\}$ (e.g., history-taking, examination recommendation, diagnosis, treatment). The patient's cognition naturally accumulates across these stages: the final cognition state of the previous task serves as the initial state of the next, namely $c_0^{(k)} = c_{T}^{(k-1)}$. Combining the within-task and cross-task updates, the cognition trajectory over the entire multi-stage pipeline is uniformly written as:
\begin{equation}
c_{t+1}^{(k)} =
\begin{cases}
\mathcal{U}\!\left(c_t^{(k)}, a_t^{(k)}, o_t^{(k)}\right) + \epsilon_t^{(k)}, & t < T_k, \\[4pt]
\mathcal{I}\!\left(c_{T_k}^{(k)}\right), & t = T_k,\; k \to k+1.
\end{cases}
\end{equation}
This design enables the patient agent to amplify the long-range impact of early decisions: a wrong early-stage decision can propagate biased cognition into later tasks, posing a more challenging and realistic environment for the doctor agent.

\subsubsection{Interactive Policy Graph-Based Doctor Agent (IPG-DA)}

To enable flexible and composable decision-making in such a dynamic environment, we introduce IPG-DA, to model the doctor agent's reasoning as a \textit{structured policy graph generation} problem rather than a fixed sequential pipeline. The doctor agent is composed of three complementary models: a \textbf{policy planner} for generating policy graphs, a \textbf{critic model} for revising them through patient interaction, and a \textbf{decision model} for executing them and producing final outputs.

\noindent\textbf{Policy Planner.}
The policy planner $\pi_\theta$ generates a DAG-structured policy graph
$G = \pi_\theta(x, c), G = (V, E, \tau)$,
where $x$ is the current task input, $c$ is the patient cognition state, $V$ is the set of nodes, $E$ is the set of dependency edges, and $\tau: V \to \mathcal{T}$ assigns each node a type from $\mathcal{T} = \{\text{interaction},\ \text{tool},\ \text{think}\}$, corresponding to doctor–patient interactions, tool invocations, and intermediate reasoning steps, respectively. During execution, the policy graph is unrolled in topological order. For each node $v \in V$, its output is computed as
$h_v = f_{\tau(v)}\!\left(\{h_u \mid u \in \text{Pa}(v)\},\ x,\ c\right)$,
where $f_{\tau(v)}$ is the operation function determined by the node type and $\text{Pa}(v)$ denotes the parent nodes of $v$. The collection $\mathcal{T}(G) = \{h_v \mid v \in V\}$ forms the complete execution trajectory.

\noindent\textbf{Critic Model.}
The critic model $\mathcal{C}_\phi$ evaluates the trajectory of the entire policy graph and identifies issues such as redundant structures, unreasonable dependencies, or missing critical information. To further verify its judgement, the critic actively interacts with the patient agent and obtains feedback
$r = \mathcal{P}(\mathcal{T}(G), c)$,
where $\mathcal{P}(\cdot)$ is the patient response function determined by cognition state $c$. Based on $G$, $\mathcal{T}(G)$ and $r$, the critic refines the policy graph through node insertion ($V \leftarrow V \cup \{v_{\text{new}}\}$), node deletion ($V \leftarrow V \setminus \{v\}$), node modification (updating $\tau(v)$ or its content), and edge revision ($E \to E'$), yielding a refined policy graph $G' = \mathcal{C}_\phi(G,\ \mathcal{T}(G),\ r)$.

\noindent\textbf{Decision Model.}
Given the refined graph $G'$, the decision model $\mathcal{F}_\psi$ aggregates the rich context produced along the execution trajectory, including symptom descriptions from interactions, returned results from tool invocations, and intermediate reasoning conclusions, into a structured medical analysis context, and produces the final task output:
$\mathcal{H}(G') = \text{Aggregate}(\mathcal{T}(G')), y = \mathcal{F}_\psi(\mathcal{H}(G'),\ x)$.

\subsubsection{Interactive Policy Graph Optimization (IPGO)}

To optimize the policy planner against the dynamic patient agent, we propose IPGO, a reinforcement learning algorithm tailored for structured policy graph generation. IPGO employs two complementary reward signals: a \textit{Policy Graph-Grounded Reward} that regularizes graph structure at the process level, and a \textit{Task-Aligned Reward} that aligns the final output with the task-specific objective.

\noindent\textbf{Policy Graph-Grounded Reward.}
This reward enforces the structural rationality of the policy graph based on node-level criteria. We first partition the nodes of $G' = (V, E, \tau)$ by type:
$V_t = \{v \in V \mid \tau(v) = t\}, t \in \mathcal{T}$.
The reward decomposes into three type-specific terms initialized as $R_{\text{int}} = R_{\text{tool}} = R_{\text{think}} = 0$, and is constrained from two aspects:

\textit{(i) Restricting Amount of Nodes.} For each node type $t$, we encourage the count $|V_t|$ to fall within a reasonable range $[N_t^{\min}, N_t^{\max}]$:
$R_t \mathrel{-}= \lambda_1 \cdot \mathbf{1}\!\left(|V_t| \notin [N_t^{\min}, N_t^{\max}]\right), t \in \mathcal{T}$,
penalizing graphs that are either too sparse or unnecessarily large.

\textit{(ii) Restricting Tool Nodes.} For each tool node $v \in V_{\text{tool}}$, we further check whether the tool exists in the tool bank, whether its inputs match the expected schema, and whether the execution is successful (no time-out or invalid output). We define
\begin{equation}
\delta(v) = 
\begin{cases}
1, & \text{if } v \text{ is illegal or fails to execute}, \\
0, & \text{otherwise},
\end{cases}
\end{equation}
and update the tool reward as
$R_{\text{tool}} \mathrel{-}= \lambda_2 \sum_{v \in V_{\text{tool}}} \delta(v)$,
where $\lambda_2$ is a penalty coefficient. The overall reward is 
$R_{\text{graph}}(G') = R_{\text{int}} + R_{\text{tool}} + R_{\text{think}}$.

\noindent\textbf{Task-Aligned Reward.}
To align the optimization with the downstream clinical task, we keep the decision model $\mathcal{F}_\psi$ {frozen} during this stage and evaluate the executed graph through the task metric:
$R_{\text{result}}(y) = \mathcal{M}(y, y^*)$,
where $y^*$ is the reference answer and $\mathcal{M}(\cdot)$ is a task-specific evaluator (e.g., accuracy or matching score). In practice, we adopt a binary form $R_{\text{result}} \in \{0, 1\}$.

\noindent\textbf{IPGO Objective.}
The total reward combines the structural and outcome signals:
$R_{\text{IPGO}} = \sum_{k \in \{\text{int},\ \text{tool},\ \text{think}\}} \lambda_k R_k + R_{\text{result}}$.
We estimate the advantage by group-wise normalization $A = R_{\text{IPGO}} - \bar{R}_{\text{IPGO}}$ and update the policy planner with the clipped surrogate loss:
\begin{equation}
\mathcal{L}_{\text{IPGO}} = \mathbb{E}\!\left[\min\!\left(rA,\ \text{clip}(r, 1 \pm \epsilon)\,A\right)\right],
\end{equation}
where $r$ denotes the importance ratio between current and old policies and $\epsilon$ is the clipping threshold. By keeping the decision model frozen, IPGO ensures that the optimization focuses on improving the {quality of generated policy graphs} rather than overfitting to a particular downstream output.

\subsection{Cross-Task Policy Graph Consolidation and Unified Model Learning (PGC-UML)}
\label{sec:stage2}

After the first stage, each task $T_k$ obtains an IPGO-optimized policy planner that produces locally optimal policy graphs. However, these graphs are tied to specific tasks and execution trajectories, and lack consistency or global optimality across the entire CDM pipeline. To address the challenge of \textit{inter-task dependencies}, we propose PGC-UML, which extracts high-value sub-structures from task-specific policy graphs, reorganizes them into a unified graph, and finally distills a unified policy planner via supervised fine-tuning.

\subsubsection{Reward-Based Policy Graph Pruning (PGP)}
To prune the graph nodes with low value, we introduce PGP to identify and retain the nodes that contribute most to the final decision for each task. Specifically, we evaluate every node by measuring its marginal effect on the final task reward. Given a policy graph $G = (V, E, \tau)$, removing a node $v \in V$ yields a pruned graph
$G^{-v} = (V \setminus \{v\},\ E^{-v},\ \tau)$,
where $E^{-v}$ is the corresponding edge set that preserves graph executability. We then re-execute both $G$ and $G^{-v}$ through the decision model:
$y = \mathcal{F}_\psi(G, x), y^{-v} = \mathcal{F}_\psi(G^{-v}, x)$,
and compute the corresponding total rewards $R(G) = R_{\text{graph}}(G) + R_{\text{out}}(y)$ and $R(G^{-v}) = R_{\text{graph}}(G^{-v}) + R_{\text{out}}(y^{-v})$. The contribution of node $v$ is defined as
$\Delta R_v = R(G) - R(G^{-v})$.
A node is regarded as a \textit{low-value node} if $\Delta R_v \approx 0$ whose removal barely affects performance, and a \textit{high-value node} if $\Delta R_v > \varepsilon$ whose removal causes a notable drop, where $\varepsilon$ is a threshold. We retain only high-value nodes, yielding a compact subgraph
$V_k^{*} \subseteq V_k$,
that preserves the core decision skeleton for task $T_k$ while discarding redundant structures.

\subsubsection{Unified Policy Graph Construction}
To capture cross-task dependencies, we further integrate high-value nodes from all tasks into a unified policy graph through three steps.

\noindent\textbf{Cross-Task Node Aggregation.}
We first aggregate high-value nodes across all $K$ tasks:
$V^{*} = \bigcup_{k=1}^{K} V_k^{*}$.
Each node in $V^{*}$ retains its type label $\tau(v) \in \mathcal{T} = \{\text{interaction},\ \text{tool},\ \text{think}\}$.

\noindent\textbf{Type-wise Node Pooling.}
Based on $\tau(\cdot)$, we partition $V^{*}$ into three type-specific node pools:
$\mathcal{P}_t = \{v \in V^{*} \mid \tau(v) = t\},\quad t \in \mathcal{T}$,
yielding $\{\mathcal{P}_{\text{interaction}}, \mathcal{P}_{\text{tool}}, \mathcal{P}_{\text{think}}\}$, which respectively correspond to the interaction node pool, tool node pool, and think node pool in Fig.~\ref{fig:main}(b).

\noindent\textbf{Inter-Node Filtering via Similarity.}
Different tasks may produce semantically or functionally similar nodes (e.g., similar questioning utterances or repeated reasoning steps). Direct aggregation would inflate the policy graph and degrade efficiency. We therefore compute pairwise cosine similarity within each pool using a node embedding $\mathbf{h}_v = \Phi(v)$:
$S_{ij} = \cos(\mathbf{h}_{v_i}, \mathbf{h}_{v_j}) = \frac{\mathbf{h}_{v_i}^\top \mathbf{h}_{v_j}}{\|\mathbf{h}_{v_i}\|\,\|\mathbf{h}_{v_j}\|}$,
and filter out highly similar nodes by a threshold $\gamma$:
$\tilde{\mathcal{P}}_t = \{v \in \mathcal{P}_t \mid \forall v' \in \tilde{\mathcal{P}}_t,\ \cos(v, v') < \gamma\},\quad t \in \mathcal{T}$.

\paragraph{Unified Policy Graph Assembly.}
After de-redundancy, we assemble the unified policy graph by selecting nodes from each pool and reconstructing edges according to their cross-task execution dependencies:
\begin{equation}
\tilde{V} = \bigcup_{t \in \mathcal{T}} \tilde{V}_t,\quad \tilde{V}_t \subseteq \tilde{\mathcal{P}}_t,\qquad \tilde{E} = \{(v_i, v_j) \mid v_i, v_j \in \tilde{V},\ v_i \prec v_j\},
\end{equation}
where $v_i \prec v_j$ denotes a topological dependency between two nodes inherited from their original task graphs. The resulting unified policy graph is $G^{*} = (\tilde{V}, \tilde{E}, \tau)$. Then, we leverage $G^{*}$ to construct a cross-task supervision dataset
$\mathcal{D}_{\text{policy}} = \{(x_i, c_i, G_i^{*}, y_i^*)\}_{i=1}^{N}$ for unified model learning.
Through this process, redundant information already covered by previous tasks is eliminated, while critical historical interactions and reasoning steps that benefit downstream decisions are explicitly preserved, leading to a compact yet globally consistent decision structure.

\subsubsection{Unified Policy Planning and Decision Learning (UPPDL)}
To model inter-task relationships and yield a globally coherent strategy, we introduce UPPDL, to learn a unified policy graph multiple tasks for the dynamic patients. Specifically, we train a \textit{unified policy planner} $\pi_\theta^{u}$ via supervised fine-tuning (SFT) on a cross-task supervision dataset $\mathcal{D}_{\text{policy}}$. To make decision model fit the output formats and evaluation criteria of each downstream task, we further fine-tune {task-specific decision models} $\{\mathcal{F}_\psi^{(k)}\}_{k=1}^{K}$ using the trajectories produced by the unified policy planner. Each decision model takes the rich context $\mathcal{H}(G^{*})$ collected during the execution of $G^{*}$ as input. This yields a unified policy planner with the specialized decision models, which jointly support consistent and globally coherent clinical decision-making across the entire CDM pipeline.

\begin{table}
\centering
\footnotesize
\setlength{\tabcolsep}{3pt}
\caption{Performance comparison on the MedChain dataset.}
\label{tab:medchain_result}
\scalebox{0.86}{
\begin{tabular}{c l c c c c c c c}
\toprule
\multirow[c]{2}{*}{\centering Framework} & 
\multirow[c]{2}{*}{\centering Methods} & 
\multicolumn{2}{c}{Specialty Referral} & 
\multirow[c]{2}{*}{\centering \makecell{History taking}} & 
\multirow[c]{2}{*}{\centering \makecell{Examination}} & 
\multirow[c]{2}{*}{\centering \makecell{Diagnosis}} & 
\multirow[c]{2}{*}{\centering \makecell{Treatment}} & 
\multirow[c]{2}{*}{\centering \makecell{Average}} \\
\cmidrule(lr){3-4}
 & & 
\centering Level1 & 
\centering Level2 & 
 &  &  &  &  \\
\midrule
\multirow{4}{*}[-0.2em]{Proprietary LLMs}
& GPT5.1~\cite{singh2026openaigpt5card}        & 0.5157 & 0.3370 & 0.5059 & 0.5638 & \underline{0.5199} & 0.5210 & 0.4939 \hspace{1pt}\diff{0.14} \\
& GPT-o3~\cite{openai_o3_2025}        & 0.5495 & 0.2866 & 0.3493 & --      & 0.4891 & 0.3343 & 0.4018 \hspace{1pt}\diff{0.23} \\
& DeepSeek V3.2~\cite{deepseekai2025deepseekv32pushingfrontieropen} & 0.5321 & 0.3530 & 0.4593 & --      & 0.5126 & 0.5769 & 0.4868 \hspace{1pt}\diff{0.14} \\
& Kimi-K2~\cite{kimiteam2026kimik2openagentic}       & 0.5227 & 0.3288 & 0.3667 & 0.4717  & 0.4297 & 0.5235 & 0.4405 \hspace{1pt}\diff{0.19} \\
\midrule
\multirow{5}{*}[-0.2em]{Open-Source LLMs}
& MedGemma-27B~\cite{sellergren2026medgemmatechnicalreport}      & 0.5571 & 0.3665 & 0.4901 & --      & 0.4879 & 0.4496 & 0.4702 \hspace{1pt}\diff{0.16} \\
& Baichuan-M2-32B~\cite{m2team2025baichuanm2scalingmedicalcapability}   & 0.5067 & 0.3337 & 0.4730 & 0.4914  & 0.5058 & 0.5882 & 0.4831 \hspace{1pt}\diff{0.15} \\
& Qwen3-32B~\cite{yang2025qwen3technicalreport}         & 0.5432 & 0.3628 & 0.5171 & 0.4819  & 0.5144 & 0.5153 & 0.4891 \hspace{1pt}\diff{0.14} \\
& Lingshu-32B~\cite{lasateam2025lingshugeneralistfoundationmodel}       & 0.5469 & 0.3673 & 0.4217 & 0.5457  & 0.4944 & 0.4934 & 0.4782 \hspace{1pt}\diff{0.15} \\
& HuatuoGPT-o1-8B~\cite{chen2024huatuogpto1medicalcomplexreasoning}  & 0.4989 & 0.3333 & 0.3402 & 0.4522  & 0.4496 & 0.2445 & 0.3865 \hspace{1pt}\diff{0.24} \\
\midrule
\multirow{4}{*}[-0.2em]{Multi-Agent System}
& MedAgent~\cite{MEDAGENTS}     & 0.3830 & 0.2039 & 0.4454 & 0.4456 & 0.4102 & 0.3673 & 0.3759 \hspace{1pt}\diff{0.26} \\
& MDAgent~\cite{mdagent}     & 0.4754 & 0.2063 & 0.4412 & 0.5346 & 0.4198 & 0.4371 & 0.4191 \hspace{1pt}\diff{0.21} \\
& MedChain-Agent~\cite{liu2025medchainbridginggapllm}   & 0.5873 & 0.3505 & \underline{0.5836} & \underline{0.6566} & 0.4807 & 0.4613 & \underline{0.5200} \hspace{1pt}\diff{0.11} \\
& MedAgent-Pro~\cite{wang2025medagentproevidencebasedmultimodalmedical} & \underline{0.5891} & \underline{0.3714} & 0.4607 & 0.5477 & 0.4520 & 0.5215 & 0.4904 \hspace{1pt}\diff{0.14} \\
\midrule
\multirow{3}{*}[-0.2em]{Self-Evolving Agent}
& MDTeamGPT~\cite{chen2025mdteamgptselfevolvingllmbasedmultiagent}     & 0.2943 & 0.3464 & 0.4446 & 0.5551 & 0.4118 & 0.4691 & 0.4202 \hspace{1pt}\diff{0.21} \\
& EvoClinician~\cite{he2026evoclinicianselfevolvingagentmultiturn}  & 0.5412 & 0.3591 & 0.3812 & 0.5226 & 0.4268 & 0.5214 & 0.4587 \hspace{1pt}\diff{0.17} \\
\rowcolor{gray!10}
& \textbf{\ours} & \textbf{0.7281} & \textbf{0.4911} & \textbf{0.5900} & \textbf{0.7248} & \textbf{0.5472} & \textbf{0.7064} & \textbf{0.6313}  \hspace{1pt} -- \textcolor{gray!10}{0.100}\\
\bottomrule
\end{tabular}
}
\end{table}

\section{Experiments}

\subsection{Datasets and Experimental Settings}
For \textbf{datasets}, we evaluate our method on two public medical CDM benchmarks: MedChain~\cite{liu2025medchainbridginggapllm} (9,724 training / 2,439 test cases) and ClinicalBench~\cite{yan2024clinicallab} (1,040 training / 460 test cases). MedChain is a multimodal benchmark and ClinicalBench is text-only benchmark. Covering both multi-stage clinical decision-making and standardized medical reasoning tasks, they enable a comprehensive evaluation across diverse clinical stages. We follow the original evaluation protocols of each benchmark for fair comparison. For \textbf{implementation details}, both the patient and doctor agents are built upon Qwen3-4B. The doctor agent includes three components: a policy planner, a critic model, and an inference model, which share the same backbone architecture. In the first stage, the decision model remains frozen and serves as a stable evaluator to provide task-aligned reward signals, while the policy planner is optimized via IPGO. The source code is available at \url{https://github.com/hotdogisme/CoE-Agent}. See Appendix Sec.~\ref{appendix_details} for details.


\subsection{Comparison with the State-of-the-Arts}
We conduct comparisons on both benchmarks against four types of approaches: {proprietary LLMs}, {open-source LLMs}, {medical multi-agent systems}, and {self-evolving agents}. As shown in Table~\ref{tab:medchain_result} and Table~\ref{tab:clinicalbench_result}, \ours~consistently surpasses all competitors with average scores of {0.6313} on MedChain and {0.7707} on ClinicalBench, outperforming the strongest baseline MedChain-Agent by {11.13\%} and {23.63\%}, respectively. Despite being built on a compact Qwen3-4B backbone, \ours~exceeds much larger proprietary and medical-specialized LLMs by 11\%–26\%, suggesting that robust interaction strategies and globally coherent policy structures matter more than parameter scaling or medical pre-training. Compared with multi-agent baselines, \ours~yields the largest gains on tasks requiring long-range cross-stage reasoning, e.g., {Specialty Referral Level1} (+13.90\%) and {Treatment} (+18.49\%) on MedChain, and {Disease Basis} (+25.46\%) and {Prescription of Treatment} (+19.34\%) on ClinicalBench, validating the benefit of modeling {inter-task dependencies} via the unified policy graph. Compared with self-evolving agents, \ours~achieves 17–21\% higher average scores, indicating that {Interactive Policy Graph Optimization} provides a more principled self-improvement signal than prompt or memory refinement. Moreover, on interaction-heavy tasks such as {History Taking} and {Examination}, where most baselines drop to 0.34–0.52 and several proprietary models even fail to produce valid outputs, \ours~still attains 0.5900 and 0.7248, confirming its robustness to {dynamic patient behaviors} and establishing a new state of the art for globally coherent clinical decision-making.

\begin{table*}[t]
\centering
\footnotesize
\setlength{\tabcolsep}{3pt}
\caption{Performance comparison on the ClinicalBench dataset.}
\label{tab:clinicalbench_result}
\resizebox{\columnwidth}{!}{
\scalebox{0.82}{
\begin{tabular}{c l cc cccccc c cccc}
\toprule

\multirow[c]{3}{*}{\centering Framework} &
\multirow[c]{3}{*}{\centering Methods} &

\multicolumn{2}{c}{Department Guide} &

\multicolumn{6}{c}{Clinical Diagnosis} &

\multirow[c]{3}{*}{\centering \makecell{Imaging\\Diagnosis}} &

\multicolumn{4}{c}{Overall} \\

\cmidrule(lr){3-4}
\cmidrule(lr){5-10}
\cmidrule(lr){12-15}

& &
Acc & DIFR &
PD & DB & DD & FD & PT & TP &
&
DWR & CDR & Acceptability & Avg. \\

\midrule

\multirow{4}{*}[-0.2em]{Proprietary LLMs}

& GPT5.1~\cite{singh2026openaigpt5card}
& 0.7000 & 1.0000
& 0.4087 & 0.3779 & 0.2853 & 0.6304 & 0.3051 & 0.2169
& 0.4023
& 8.2609 & 0.5870 & 0.2233
& 0.4807 \hspace{1pt}\diff{0.29} \\

& GPT-o3~\cite{openai_o3_2025}
& 0.7391 & \underline{0.9993}
& 0.4022 & 0.4213 & 0.3548 & 0.6065 & 0.3504 & 0.2290
& 0.3947
& 10.6087 & 0.5565 & 0.2256
& 0.4997 \hspace{1pt}\diff{0.27} \\

& DeepSeek V3.2~\cite{deepseekai2025deepseekv32pushingfrontieropen}
& 0.7043 & 0.9926
& 0.4587 & 0.4508 & 0.3432 & \underline{0.6370} & 0.3086 & 0.2780
& 0.5041
& 11.4783 & 0.5739 & 0.2477
& 0.5197 \hspace{1pt}\diff{0.25} \\

& Kimi K2~\cite{kimiteam2026kimik2openagentic}
& 0.5804 & 0.9133
& 0.3022 & 0.3763 & 0.2500 & 0.5217 & 0.3428 & 0.3599
& 0.4075
& 5.3913 & 0.4543 & 0.1822
& 0.4504 \hspace{1pt}\diff{0.32} \\

\midrule

\multirow{5}{*}[-0.2em]{Open-Source LLMs}

& MedGemma-27B~\cite{sellergren2026medgemmatechnicalreport}
& 0.6826 & 0.9847
& 0.3261 & 0.4597 & 0.3337 & 0.5587 & 0.3692 & 0.2851
& 0.3411
& 6.5652 & 0.4783 & 0.2002
& 0.4823 \hspace{1pt}\diff{0.30} \\

& Baichuan-M2-32B~\cite{m2team2025baichuanm2scalingmedicalcapability}
& 0.6978 & 0.9953
& 0.4161 & \underline{0.4628} & 0.3482 & 0.5643 & 0.3157 & 0.2656
& 0.4156
& 10.5652 & 0.5543 & 0.2274
& 0.4979 \hspace{1pt}\diff{0.27} \\

& Qwen3-32B~\cite{yang2025qwen3technicalreport}
& 0.6108 & 1.0000
& 0.3543 & 0.4379 & 0.3273 & 0.4848 & 0.3282 & 0.2579
& 0.4166
& 7.2609 & 0.4978 & 0.2034
& 0.4687 \hspace{1pt}\diff{0.30} \\

& Lingshu-32B~\cite{lasateam2025lingshugeneralistfoundationmodel}
& 0.6695 & 0.9820
& 0.3870 & 0.4018 & \underline{0.4289} & 0.5522 & \underline{0.3873} & \underline{0.4069}
& 0.3795
& 13.2174 & 0.5717 & 0.2626
& 0.5105 \hspace{1pt}\diff{0.26} \\

& HuatuoGPT-o1-8B~\cite{chen2024huatuogpto1medicalcomplexreasoning}
& 0.5173 & 0.9960
& 0.1369 & 0.1929 & 0.1887 & 0.3344 & 0.2218 & 0.3891
& 0.1190
& 1.0000 & 0.2543 & 0.0689
& 0.3440 \hspace{1pt}\diff{0.43} \\

\midrule

\multirow{4}{*}[-0.2em]{Multi-Agent System}

& MedAgent~\cite{MEDAGENTS}
& 0.6760 & 0.9935
& 0.3122 & 0.4371 & 0.3394 & 0.4163 & 0.2801 & 0.2815
& 0.4195
& 5.3478 & 0.4130 & 0.1664
& 0.4617 \hspace{1pt}\diff{0.31} \\

& MDAgent~\cite{mdagent}
& 0.6543 & 0.9978
& 0.3311 & 0.4259 & 0.3347 & 0.4561 & 0.3171 & 0.2835
& 0.3369
& 4.3043 & 0.4783 & 0.1853
& 0.4597 \hspace{1pt}\diff{0.31} \\

& MedChain-Agent~\cite{liu2025medchainbridginggapllm}
& 0.8239 & 0.9978
& 0.4783 & 0.4421 & 0.3423 & 0.6000 & 0.3541 & 0.2666
& \underline{0.5043}
& \underline{13.9565} & 0.6065 & \underline{0.2645}
& \underline{0.5344} \hspace{1pt}\diff{0.24} \\

& MedAgent-Pro~\cite{wang2025medagentproevidencebasedmultimodalmedical}
& \underline{0.8586} & 0.9870
& \underline{0.5065} & 0.3159 & 0.2761 & 0.5935 & 0.3195 & 0.2687
& 0.2997
& 6.0435 & 0.5717 & 0.1892
& 0.4917 \hspace{1pt}\diff{0.28} \\

\midrule

\multirow{3}{*}[-0.2em]{Self-Evolving Agent}

& MDTeamGPT~\cite{chen2025mdteamgptselfevolvingllmbasedmultiagent}
& 0.7260 & 0.9935
& 0.3457 & 0.4346 & 0.3321 & 0.4978 & 0.3422 & 0.2659
& 0.3454
& 7.1304 & 0.5391 & 0.2084
& 0.4759 \hspace{1pt}\diff{0.29} \\

& EvoClinician~\cite{he2026evoclinicianselfevolvingagentmultiturn}
& 0.6413 & 1.0000
& 0.4370 & 0.4285 & 0.3225 & \underline{0.6370} & 0.3247 & 0.2591
& 0.3922
& 9.3913 & \underline{0.6130} & 0.2457
& 0.4936 \hspace{1pt}\diff{0.28} \\

\rowcolor{gray!10}
& \textbf{\ours}
& \textbf{0.8717} & \textbf{1.0000}
& \textbf{0.5369} & \textbf{0.9802} & \textbf{0.5966} & \textbf{0.7241} & \textbf{0.8514} & \textbf{0.6752}
& \textbf{0.7001}
& \textbf{15.4783} & \textbf{0.6391} & \textbf{0.4058}
& \textbf{0.7707} \hspace{1pt} -- \textcolor{gray!10}{0.100}\\

\bottomrule
\end{tabular}
}}
\end{table*}

\subsection{Ablation Study}
To assess the effectiveness of each component in \ours, we conduct an ablation study on the MedChain benchmark by progressively enabling the Dynamic Patient Agent (DPA), the Interactive Policy Graph-Based Doctor Agent (IPG-DA) within Co-Evolutionary Learning (CoEL), and the Cross-Task Policy Graph Consolidation and Unified Model Learning (PGC-UML) in Table~\ref{tab:ablation}. Starting from the baseline (0.4465), adding DPA alone improves the average to 0.5528, indicating that simulating non-stationary and unreliable patient behaviors fosters more robust decision-making. Incorporating IPG-DA further enables structured policy graph generation with critic-driven revision, yielding additional gains over fixed sequential pipelines (0.5220). Applying PGC-UML alone raises performance to 0.5500, confirming that modeling inter-task dependencies via node pruning, cross-task aggregation, and unified planner distillation is essential for globally coherent decision-making. Finally, combining all three components achieves the best result of 0.6313 (+18.48\% over the baseline), showing that the modules are highly complementary. 

\begin{table}[t]
\centering
\begin{minipage}[t]{0.47\textwidth}
    \centering
    \footnotesize
    \caption{Ablation study on the MedChain dataset.}
    \label{tab:ablation}
    \begin{tabular}{c c c c} 
    \toprule
    \multicolumn{2}{c}{CoEL} &
    \multirow{2}{*}{PGC-UML} & 
    \multirow{2}{*}{Average} \\
    \cmidrule(lr){1-2} 
     DPA & IPG-DA &  & \\ 
    \midrule
     &  &  & 0.4465 \\
    $\checkmark$ &  &  & 0.5528 \\
     & $\checkmark$ &  & 0.5220 \\
     &  & $\checkmark$ & 0.5500 \\
    $\checkmark$ & $\checkmark$ & $\checkmark$ & \textbf{0.6313} \\
    \bottomrule
    \end{tabular}
\end{minipage}
\hfill 
\begin{minipage}[t]{0.5\textwidth}
    \centering
    \footnotesize
    \caption{Computational efficiency Comparison.}
    \label{tab:efficiency}
    \begin{tabular}{l l r}
    \toprule
    {Framework} & {Method} & {Inference Time / s} \\
    \midrule
    \multirow{4}{*}{\makecell[c]{Multi-Agent\\System}} 
    & MedAgent~\cite{MEDAGENTS}      & 98.3 \\
    & MDAgent~\cite{mdagent}       & 162.6 \\
    & MedChain~\cite{liu2025medchainbridginggapllm}      & 94.6 \\
    & MedAgent-Pro~\cite{wang2025medagentproevidencebasedmultimodalmedical}  & 154.8 \\
    \midrule
    \multirow{3}{*}{\makecell[c]{Self-Evolving\\Agent}} 
    & MDTeamGPT~\cite{chen2025mdteamgptselfevolvingllmbasedmultiagent}     & 341.2 \\
    & EvoClinician~\cite{he2026evoclinicianselfevolvingagentmultiturn}  & 180.7\\
    & \textbf{\ours} & \textbf{17.5} \\
    \bottomrule
    \end{tabular}
\end{minipage}
\end{table}

\begin{figure*}[t]
	\centering
	\includegraphics[width=\textwidth]{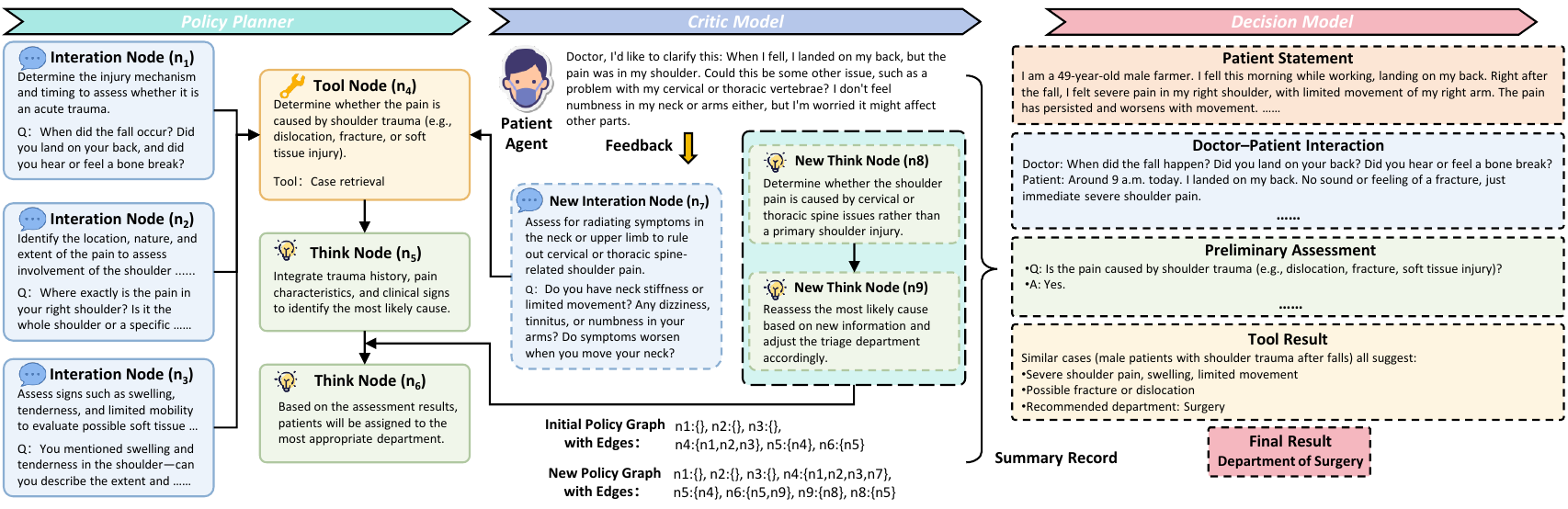}	
	\caption{Visualization of Interactive Policy Graph-Based Doctor Agent.}
	\label{fig:doctor_agent_visual_analysis}
\end{figure*}

\subsection{Computational Efficiency}
Beyond accuracy, inference efficiency is critical for deploying medical agents in real clinical settings. We compare the average per-case inference time of \ours~with seven baselines on MedChain. In Table~\ref{tab:efficiency}, \ours~requires only {17.5s per case}, achieving a $5.4\times$–$9.3\times$ speedup over multi-agent systems and up to $19.5\times$ speedup over self-evolving agents. This efficiency mainly stems from the unified policy planner that replaces costly multi-role coordination with a single planning pass, together with reward-based pruning that removes redundant nodes to yield compact policy graphs.

\subsection{Visualization of Interactive Policy Graph-Based Doctor Agent }

To visualize the interpretability of doctor agent, Fig.~\ref{fig:doctor_agent_visual_analysis} visualizes case about right scapular fracture from MedChain, walking through the three components of the doctor agent, i.e. policy planner, critic model, and decision model. Given a first-person chief complaint from the patient agent, policy planner emits a 6-node DAG-structured {initial policy graph} comprising three node types, each defined by an explicit {purpose} and a type-specific operation: {interaction} nodes ($n_1$–$n_3$) actively elicit information via targeted queries (e.g., \textit{``When did the fall occur? Did you land on your back?''}); the {tool} node ($n_4$) invokes an external tool to verify the trauma hypothesis; and {think} nodes ($n_5, n_6$) perform internal sub-reasoning over upstream outputs to integrate evidence toward triage. After execution, critic model collects patient feedback and supports three revision operations (e.g. {add}, {delete}, {modify}) of which this case illustrates {add}: when the patient questions a possible \textit{``cervical or thoracic''} origin, critic model inserts an interaction node $n_7$ (\textit{``Assess radiating symptoms''}) and think nodes $n_8, n_9$ (\textit{``Rule out spinal origin''}), and reconstructs edges to preserve DAG validity, yielding the {new policy graph} ($n_4{:}\{n_1,n_2,n_3,n_7\}$, $n_6{:}\{n_5,n_9\}$, $n_9{:}\{n_8\}$, $n_8{:}\{n_5\}$). Finally, decision model aggregates the refined trajectory into four structured fields, and outputs \emph{Department of Surgery}, consistent with the ground truth, jointly verifying that policy model produces semantically complete DAGs, critic model converts patient feedback into valid structural edits, and decision model derives globally coherent clinical conclusions.

\section{Conclusion}
In this paper, we presented CoE-Agent, a co-evolving patient–doctor agentic framework for CDM. It integrates a CoEL stage, where a DPA with an evolving cognition state interacts with an IPG-DA optimized via IPGO for structural validity and task alignment, and a PGC-UML stage that prunes, aggregates, and distills task-specific policy graphs into a unified planner for globally coherent decision-making. Extensive experiments show that \ours~significantly outperforms current baselines with substantial inference speedups. While \ours~surpasses much larger proprietary LLMs on a single Qwen3-4B backbone, its behavior at larger scales (e.g., 32B or 70B) or across heterogeneous backbones remains future work. Broadly, \ours~has the potential to assist medical education, support clinical reasoning training, and alleviate diagnostic workloads in under-resourced settings through interpretable, policy-guided decision-making.


\begin{ack}
This work was supported by the National Natural Science Foundation of China under Grant 62576216, Guangdong Provincial Key Laboratory under Grant 2023B1212060076, and supported by the Intelligent Computing Center of Shenzhen University.
\end{ack}




\bibliography{custom}
\bibliographystyle{plain}






\medskip
\clearpage

\appendix

\noindent { \LARGE \textbf{Appendix for \ours}}

\smallskip
\smallskip
\smallskip

\noindent\textbf{Abstract.} 

Appendix A presents the comprehensive ablation study on the MedChain dataset.

Appendix B presents the computational efficiency analysis, comparing our method with medical multi-agent systems and self-evolving approaches, and demonstrating the advantage of our unified policy mechanism in reducing inference cost.

Appendix C provides a detailed visualization of the Patient and Doctor Agents, including the construction of the dynamic cognition state, its update process, and how error propagation occurs across sequential clinical tasks.

Appendix D illustrates the policy graph for each task and the unified policy graph, highlighting how high-value decision structures are extracted, merged, and reused across stages to achieve global policy optimization.

Appendix E provides the datasets and implementation details.

Appendix F and G include the prompts used for the Patient Agent, Doctor Agent, and Unified Policy Planner, respectively, providing additional implementation details for reproducibility.

\section{Ablation Study}

To validate the effectiveness of each component, we conduct systematic ablation studies on the key modules of our framework, as shown in Table~\ref{tab_appendix:ablation}. Overall, our method consists of three core components: the Dynamic Patient Agent, the Interactive Policy Graph-Based Doctor Agent, and Cross-Task Policy Graph Consolidation and Unified Model Learning. Notably, the latter is tightly coupled with the Doctor Agent; when used independently, it degenerates into standard SFT on the Decision Model.

Starting from the base setting without any modules (i.e., using the raw model with static inputs), the overall performance is relatively low (0.4465), indicating that direct decision-making based on static inputs is insufficient for complex clinical workflows. When only patient dialogue data is introduced (i.e., replacing raw formatted input with policy graph representation), performance improves to 0.5528, demonstrating that policy graph modeling is highly effective. Even without optimization, using policy graphs alone yields a 10.63\% improvement.

Further, applying SFT alone (without policy graph modeling) results in a performance of 0.5500, showing limited gains and suggesting that static supervised learning is insufficient for modeling complex decision processes. When combining policy graph modeling with SFT, performance increases to 0.5881, achieving an additional 3.81\% improvement, which indicates that structured policy representation is more effective than static structured outputs. When both the Patient Agent and Doctor Agent are introduced, performance further improves significantly to 0.6288, validating the effectiveness of combining structured policy modeling with reinforcement learning optimization.

Finally, integrating all components yields the best performance (0.6313). Compared to the setting without Cross-Task Policy Graph Consolidation and Unified Model Learning, performance improves by 0.25\%. This result shows that cross-task policy consolidation can effectively integrate high-value decision structures across tasks, enabling global policy optimization and further improving performance in multi-stage clinical workflows. At the same time, it unifies multi-stage policy graphs into a single model, significantly reducing computational cost and resource consumption while maintaining strong performance.


\begin{table}
\centering
\footnotesize
\setlength{\tabcolsep}{4pt}
\caption{Ablation study on the MedChain dataset.}
\label{tab_appendix:ablation}
\resizebox{\columnwidth}{!}{
\begin{tabular}{c c c c c c c c c c}
\toprule
\multicolumn{2}{c}{CoEL} &
\multirow{2}{*}{PGC-UML} & 
\multicolumn{2}{c}{Specialty Referral} & 
\multirow{2}{*}{History} & 
\multirow{2}{*}{Exam} & 
\multirow{2}{*}{Diag} & 
\multirow{2}{*}{Treat} & 
\multirow{2}{*}{Avg} \\
\cmidrule(lr){4-5} \cmidrule(lr){1-2} 
 DPA & IPG-DA & & Level1 & Level2 & & & & & \\
\midrule

 &  &  & 0.5153 & 0.3234 & 0.4441 & 0.4734 & 0.4750 & 0.4479 & 0.4465 \\

$\checkmark$ &  &  & 0.6478 & 0.4225 & 0.4392 & 0.6658 & 0.5050 & 0.6367 & 0.5528 \\
 
 & $\checkmark$ &  & 0.6022 & 0.4610 & 0.3863 & 0.5537 & 0.5137 & 0.6153 & 0.5220 \\

 &  & $\checkmark$ & 0.6929 & 0.4634 & 0.5452 & 0.4808 & 0.4665 & 0.6513 & 0.5500 \\



$\checkmark$ & $\checkmark$ & $\checkmark$ 
& \textbf{0.7281} & \textbf{0.4911} & \textbf{0.5900} & \textbf{0.7248} & \textbf{0.5472} & \textbf{0.7064} & \textbf{0.6313} \\

\bottomrule
\end{tabular}
}
\end{table}

\section{Computational Efficiency}

We further evaluate the computational efficiency of different methods, as shown in Table~\ref{tab_appendix:efficiency}. The compared approaches mainly include medical multi-agent systems (MMAS) and self-evolving methods. The reported results correspond to the average inference time for the Specialty Referral task on the MedChain dataset. Since a complete clinical workflow in MedChain consists of five sequential tasks, the overall inference cost of these methods can be approximated as single-task time × 5. Methods such as MedAgent and MedChain rely on multi-agent collaboration to perform stage-wise decision-making, while MDAgent and MedAgent-Pro further introduce complex coordination and tool usage. Self-evolving approaches (e.g., MDTeamGPT and EvoClinician) additionally require iterative optimization and experience accumulation, leading to even higher inference costs.

In contrast, our method demonstrates a clear advantage in computational efficiency. For a single task, our approach achieves the lowest inference time, requiring only 17.45 seconds. This indicates that although our framework involves a relatively complex policy graph, its actual execution cost remains low while still delivering significant performance gains, validating the effectiveness of our policy graph modeling. More importantly, our method completes the entire clinical workflow in 17.45 seconds (×1), whereas other methods require repeated execution across multiple stages (e.g., MedAgent: 19.66 × 5, MDAgent: 32.51 × 5). Overall, our method achieves substantially lower inference time compared to all baselines.

This efficiency gain mainly stems from our proposed Cross-Task Policy Graph Consolidation and Unified Model Learning mechanism. Through this module, we integrate high-quality decision structures across tasks and learn a unified global policy. During inference, this unified policy models the entire preceding decision process in a single pass, eliminating the need to repeatedly execute full multi-agent interactions or self-evolving procedures for each subtask. Subsequently, the final outputs for each stage are obtained by simply parsing the execution results using the Decision Model. Experimental results demonstrate that our method not only outperforms existing approaches in performance but also achieves superior computational efficiency. This “generate once, reuse across tasks” paradigm significantly enhances practicality and scalability for real-world clinical applications.


\begin{table}[t]
\centering
\caption{Computational Efficiency Comparison (Inference Time in Seconds)}
\label{tab_appendix:efficiency}
\begin{tabular}{l l c}
\toprule
\textbf{Framework} & \textbf{Method} & \textbf{Inference Time / s} \\
\midrule
\multirow{4}{*}{MAS} 
& MedAgent      & 19.66 * 5\\
& MDAgent       & 32.51 * 5 \\
& MedChain      & 18.92 * 5 \\
& MedAgent-Pro  & 30.96 * 5 \\
\midrule
\multirow{2}{*}{Self-evolution} 
& MDTeamGPT     & 68.24 * 5 \\
& EvoClinician  & 36.16 * 5 \\
\midrule
& \textbf{Ours} & \textbf{17.45  * 1} \\
\bottomrule
\end{tabular}
\end{table}

\begin{figure*}[t]
	\centering
	\includegraphics[width=\textwidth]{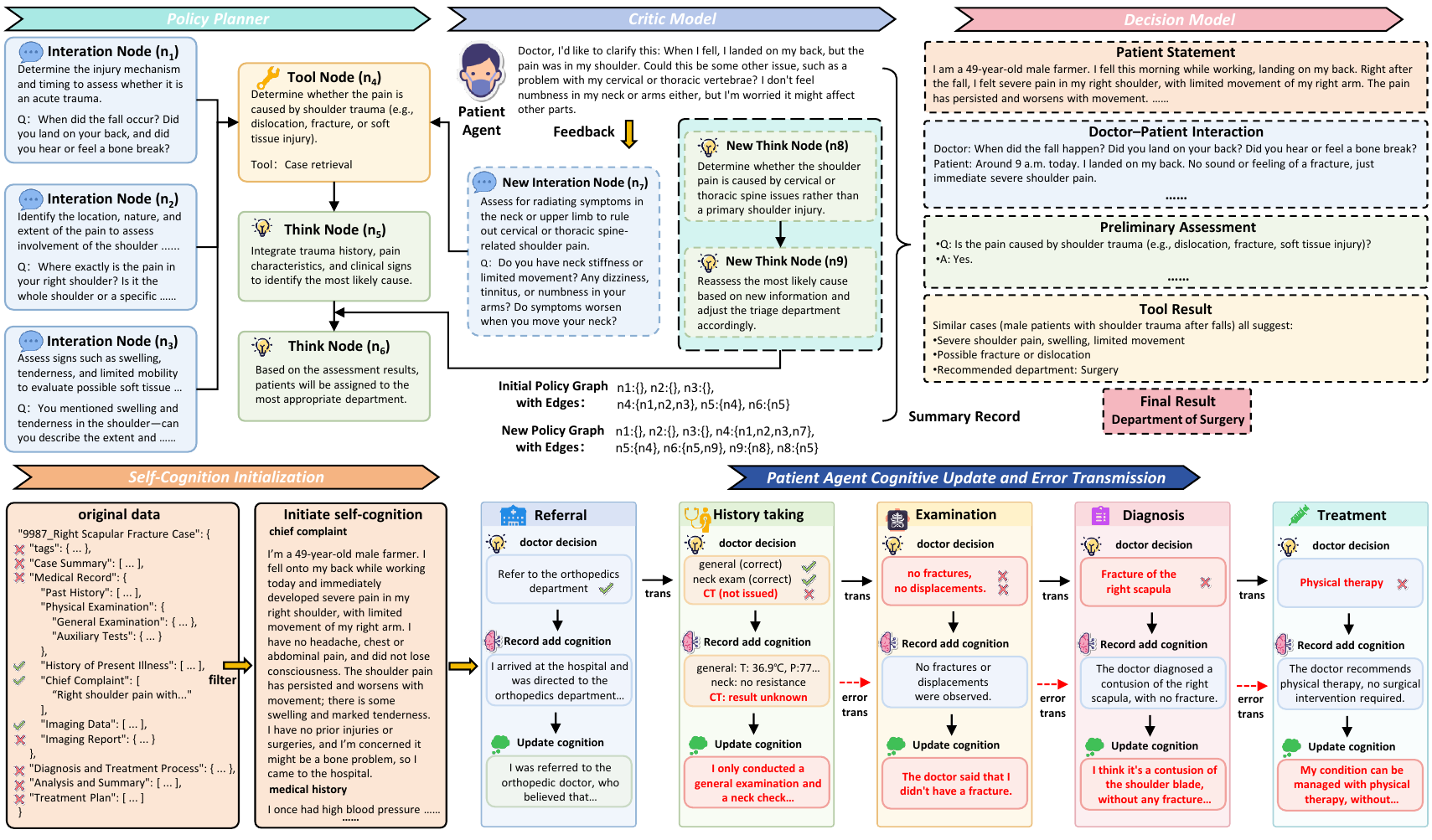}	
	\caption{Visual analysis of doctor agent in triage task of the MedChain dataset.}
	\label{fig_appendix:patient_agent_visual_analysis}
\end{figure*}

\section{Visualization of Patient and Doctor Agents}

The core of the Patient Agent lies in a dynamically updated cognition state, which serves as a unified representation of all information the patient “perceives” at a given moment, including symptom descriptions, medical history, examination results, and the patient’s understanding of their own condition. Unlike traditional approaches that model the patient as a static information provider, we treat it as a continuously evolving cognitive entity driven by interaction. The overall construction of the Patient Agent, along with the processes of cognition updating and information propagation, is illustrated in Figure~\ref{fig_appendix:patient_agent_visual_analysis}.

At the initialization stage, the cognition state is constructed solely from information available at the initial consultation, such as the chief complaint and basic medical history. All future information (e.g., examination results and final diagnoses) is strictly removed to prevent information leakage. The filtered patient information is then reformulated in the first-person perspective and provided to the Patient Agent for role-playing, closely simulating real clinical scenarios. This primarily includes the current chief complaint and past medical history, enabling the Patient Agent to form an initial understanding of its condition and support subsequent interactions.

In the multi-stage clinical workflow, the cognition state is continuously propagated and accumulated across tasks, forming a cross-stage trajectory of cognitive evolution. This design allows early decisions to exert long-term effects on subsequent tasks, explicitly modeling error propagation and information dependency in real-world clinical settings. Specifically, the doctor’s questions, examination recommendations, and diagnostic conclusions act as external inputs that influence and update the patient’s cognition. These updates not only incorporate newly acquired information (e.g., examination results) but may also adjust or even distort existing cognition. For instance, when the doctor makes an incorrect judgment, the patient may develop a misunderstanding, leading to inconsistent or misleading responses in later interactions. In contrast, accurate clinical decisions help progressively refine the patient’s cognition, resulting in more stable and reliable information expression. As shown in Fig.~\ref{fig_appendix:patient_agent_visual_analysis}, we present an example from the MedChain dataset (“Case 9987: Right Scapular Fracture”) to illustrate cognition initialization, updating, and error propagation. In this case, during the Specialty Referral stage, the Doctor Agent makes correct decisions, and the execution results of the policy graph are incorporated into the Patient Agent’s cognition and carried forward to subsequent tasks. However, in the History Taking stage, if the Doctor Agent orders incorrect or incomplete examinations (e.g., missing a CT scan), the corresponding results cannot be retrieved, leading to errors that are incorporated into the cognition state and subsequently affect downstream tasks. More critically, if the Doctor Agent makes an incorrect diagnosis or description at a key stage, it can cause catastrophic failure in all subsequent tasks. For example, in the Examination stage of this case, the Doctor Agent incorrectly interprets the imaging results as showing no fracture or displacement. This incorrect information is then integrated into the Patient Agent’s cognition, which directly leads to an incorrect diagnosis in the Diagnosis stage (e.g., diagnosing a shoulder contusion instead of a fracture), and consequently results in an inappropriate treatment plan (e.g., choosing physical therapy instead of surgery). Compared to MedChain-Agent, the error propagation in our setting is not limited to a single subsequent input but persists throughout all following interaction stages.

From an interpretability perspective, the behavior of the Patient Agent can be viewed as an external manifestation of its cognition state: each response is generated based on the current cognition, and changes in cognition directly reflect the flow of information and the impact of decisions during doctor–patient interactions. Therefore, this design not only enhances the realism of the simulated environment but also provides a clear and traceable explanation pathway for analyzing model behavior in complex clinical interactions. 


\section{Visualization of Policy Graph for Each Task and Unified Policy Graph}

\begin{figure*}[t]
	\centering
	\includegraphics[width=\textwidth]{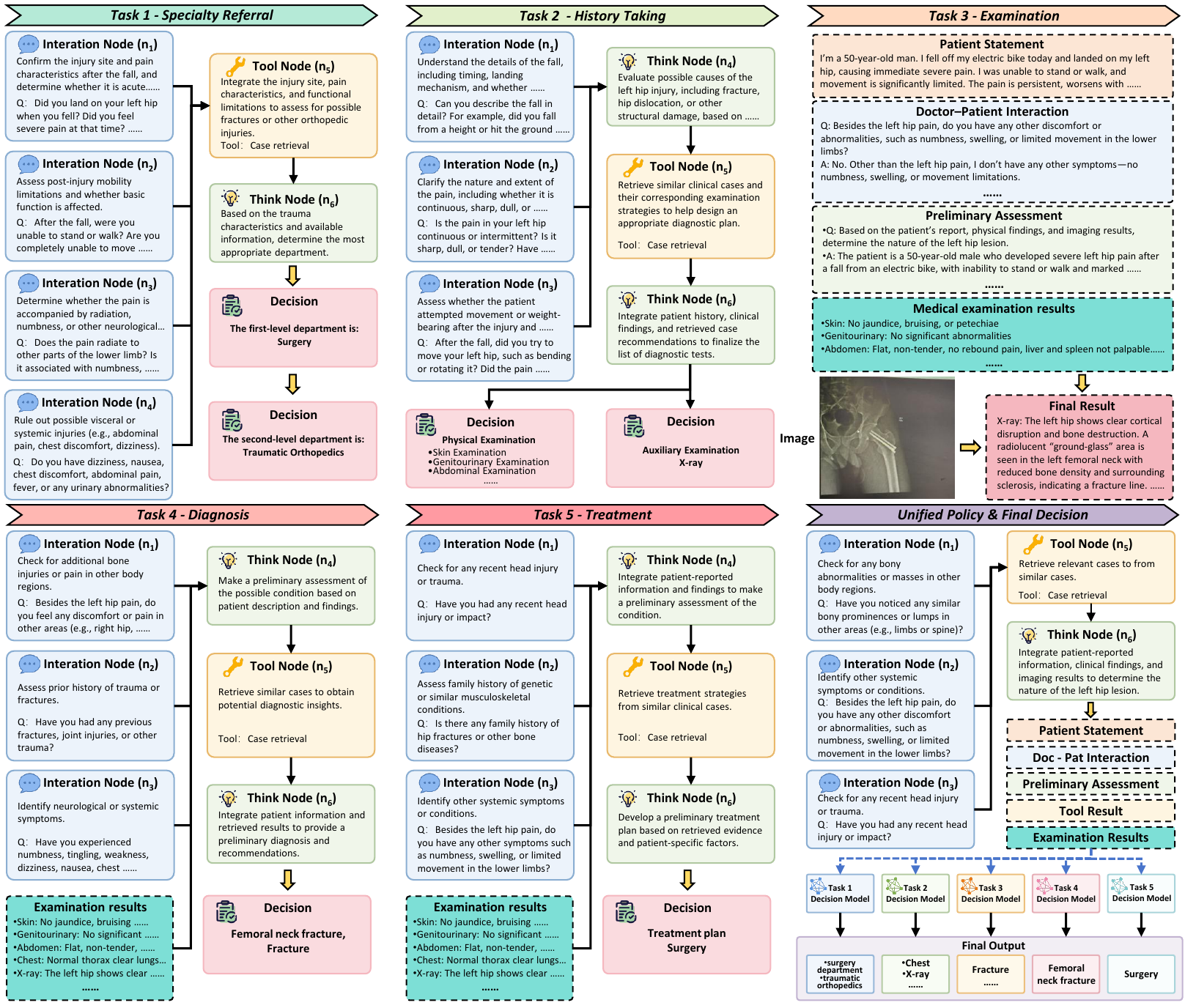}	
	\caption{Visual analysis of doctor agent across all the tasks from the MedChain dataset.}
	\label{fig_appendix:unified policy graph}
\end{figure*}

To intuitively illustrate the behavior of the proposed method in multi-stage clinical decision-making, we conduct a comprehensive visualization analysis of the policy graphs generated for each subtask (Task 1–Task 5) as well as the final unified policy graph. A representative case from the MedChain dataset, “Case 9974: Left Femoral Neck Fracture,” is selected for detailed analysis, as shown in Figure~\ref{fig_appendix:unified policy graph}. For clarity, this case study focuses on the construction of the policy graph at each stage and the unified policy graph, while omitting the Critic Model process.

In the Specialty Referral stage (Task 1), the policy graph primarily revolves around the patient’s initial complaint. The graph consists of three types of nodes. The Policy Planner first generates a complete policy graph based on the patient’s statement and prior context, and the execution results are then passed to the Decision Model for final prediction. Typically, the graph begins with interaction nodes, which acquire key clinical information—such as injury mechanism, pain location, and severity—through multi-turn questioning (e.g., “Confirm the injury site and pain characteristics after the fall, and determine whether it is acute…”, “Assess post-injury mobility limitations and whether basic function is affected.”). In this setting, all relevant information must be actively obtained through doctor–patient interaction (e.g., “Q: Do you have dizziness, nausea, chest discomfort, abdominal pain, fever, or any urinary abnormalities?”). After sufficient interaction, the model obtains critical clinical insights. These interaction nodes are followed by either tool nodes or think nodes. Tool nodes specify the use of external tools and return corresponding results, while think nodes perform intermediate reasoning based on previously collected information, aiming to deepen reasoning and rule out alternative conditions. After all nodes are executed, the aggregated information is fed into the Decision Model to produce the final output. Importantly, the decisions made by the Doctor Agent at each stage directly influence the Patient Agent’s cognition and the overall task outcome, thereby affecting the accuracy of information available in subsequent stages. For instance, if incorrect examinations are prescribed in the History Taking stage (Task 2) or misinterpretations occur in the Examination stage (Task 3), such errors (as illustrated in the “Medical examination results” in Figure~\ref{fig_appendix:unified policy graph}) will propagate to all subsequent tasks, leading to compounded failures.

Within the Cross-Task Policy Graph Consolidation and Unified Model Learning framework, we further construct a unified policy planner that generates a single policy graph covering the entire five-stage clinical workflow. Compared to generating separate policy graphs for each task, the unified graph is more compact in structure, often with fewer or comparable nodes but significantly higher information density. This is achieved by retaining only high-value decision nodes during the consolidation process and merging semantically or functionally redundant nodes across tasks, thereby reducing redundancy. Moreover, the unified policy graph explicitly models cross-stage dependencies, allowing key decision nodes to serve multiple stages without repeated generation.

Specifically, using interaction nodes as examples, we analyze the correspondence between the unified policy and task-specific policies to validate its effectiveness. The interaction nodes in the unified policy can be clearly traced back to multiple task stages, demonstrating that the unified policy is not newly generated but rather a compression and reorganization of high-value decision patterns. For instance, the unified node “Check for any recent head injury or trauma” can be traced to core inquiries in Task 1 and Task 2 regarding injury mechanisms. In Task 2, this corresponds to “Understand the details of the fall, including timing and landing mechanism,” while in Task 1 it relates to determining whether the injury is acute. These nodes repeatedly appear across tasks, highlighting their fundamental importance in clinical reasoning and justifying their inclusion in the unified policy.

Similarly, the unified node “Identify other systemic symptoms or conditions” also has cross-task origins. In Task 1, it corresponds to “Rule out possible visceral or systemic injuries,” serving as an initial risk screening step, while in Task 4 it appears as “Identify neurological or systemic symptoms” to support diagnosis and differential analysis. This demonstrates that the node spans multiple stages from early screening to diagnosis, and is therefore abstracted into a more general form in the unified policy. Finally, the node “Check for any bony abnormalities or masses in other body regions” closely aligns with “Check for additional bone injuries or pain in other body regions” in Task 4, serving as an extension of diagnostic reasoning to avoid missing multiple injuries or related conditions. These observations show that nodes in the unified policy are not isolated but originate from repeatedly occurring, high-impact decision patterns across tasks. By abstracting and merging these nodes, the unified policy preserves essential cross-task decision logic while significantly reducing redundancy, thereby improving both efficiency and generalization without sacrificing performance.

\section{Datasets and Implementation Details}
\label{appendix_details}
\paragraph{Datasets and Evaluation Metrics.} We evaluate the proposed method on two public medical benchmarks: MedChain and ClinicalBench. These datasets cover both multi-stage clinical decision-making processes and standardized medical reasoning tasks, enabling a comprehensive assessment of model performance across different clinical stages. Importantly, they exhibit strong sequential dependency, where decisions made in earlier stages directly influence the availability of information and decision pathways in subsequent stages, making them well-suited for evaluating cross-stage dependency modeling and global policy optimization. To ensure fair comparison, we follow prior work and adopt their evaluation protocols. Notably, we use the Chinese version of the datasets. In ClinicalBench, we evaluate department guide task via accuracy and Department Instruction Following Rate (DIFR). In clinical diagnosis, PD task is assessed via accuracy, and the rest of task is evaluated through the average results of BLEU, ROUGE, and BERTScore. For the overall assessment, we compute the average (Avg.) score over all the 9 scores under the 8 tasks, and use Department Win Rate (DWR) and Comprehensive Diagnostic Rate (CDR) to evaluate.

\paragraph{Implementation Details.} For MedChain, we split the dataset into training and test sets with a ratio of 8:2. For ClinicalBench, the training set contains 1,040 samples and the test set contains 460 samples. Both the Patient Agent and Doctor Agent are built upon Qwen3-4B. The Doctor Agent consists of three core components: a policy model, a critic model, and an inference model, which share the same backbone architecture but serve different roles. For training, our proposed Interactive Policy Graph Optimization (IPGO) is implemented based on Group Relative Policy Optimization (GRPO) to optimize the policy graph generated by the Policy Planner. During this process, the Decision Model remains frozen and serves as a stable evaluator to provide outcome-aligned reward signals, while the Policy Planner is updated under both structural constraints and task feedback. In the second stage, we perform supervised fine-tuning (SFT) on the Unified Policy Planner using the recomposed high-quality policy graph data. Furthermore, we leverage the execution trajectories of the policy graphs to conduct task-aligned training of the Decision Model, improving its performance on downstream tasks. 
The hyperparameters of IPGO are configured as follows. We adopt a learning rate of $5\times10^{-6}$ with the AdamW optimizer ($\beta_1=0.9$, $\beta_2=0.99$, weight decay $=0.1$), and employ a cosine learning rate scheduler with a warmup ratio of $0.1$. The model is trained using bfloat16 precision with a per-device batch size of $4$ and gradient accumulation steps of $4$. During training, the maximum prompt length and completion length are both set to $1024$, and the number of generations for GRPO sampling is set to $4$. Gradient clipping is applied with a maximum gradient norm of $0.1$. For parameter-efficient fine-tuning, we employ LoRA with rank $r=32$, LoRA alpha $=64$, and dropout $=0.05$. The LoRA modules are applied to the attention and feed-forward layers, including \texttt{q\_proj}, \texttt{k\_proj}, \texttt{v\_proj}, \texttt{o\_proj}, \texttt{gate\_proj}, \texttt{up\_proj}, and \texttt{down\_proj}. All experiments are based on the Qwen3-4B model initialized in bfloat16 format. We use DeepSpeed ZeRO-3 offloading for memory optimization during training. All experiments are conducted on $4$ NVIDIA A6000 GPUs, and the entire training process is completed in less than two days.

\section{Prompts for Patient Agent}
\begin{figure*}[ht]
	\centering
	\includegraphics[width=\textwidth]{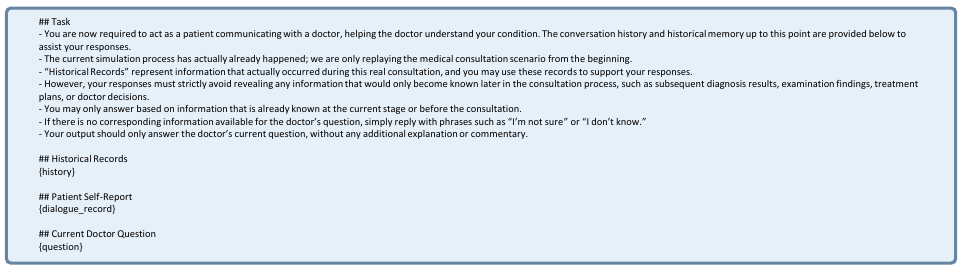}	
	\caption{The prompt of the patient's responses during the doctor-patient interaction.}
	\label{fig_appendix:patient_answer}
\end{figure*}

\begin{figure*}[t]
	\centering
	\includegraphics[width=\textwidth]{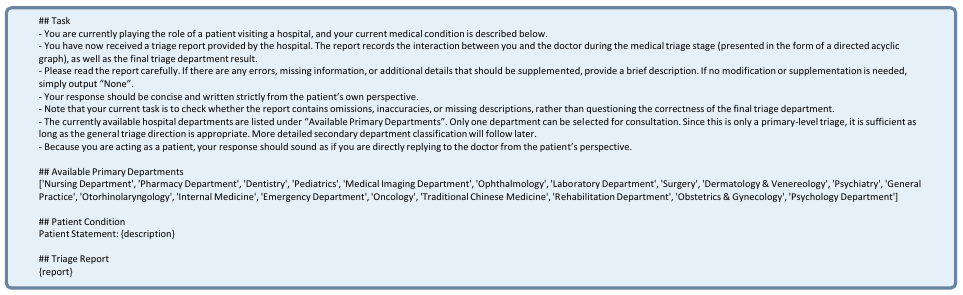}	
	\caption{The prompt of the patient's opinion on the policy.}
	\label{fig_appendix:patient_view}
\end{figure*}

In this appendix, we further provide the detailed prompts used for the construction of the Patient Agent. Specifically, we present the prompts for two core components: (1) the patient response generation process during doctor--patient interaction, and (2) the patient feedback generation process for evaluating and commenting on the triage report. As shown in Fig.~\ref{fig_appendix:patient_answer} and Fig.~\ref{fig_appendix:patient_view}

\section{Prompts for Doctor Agent}
\begin{figure*}[t]
	\centering
	\includegraphics[width=\textwidth]{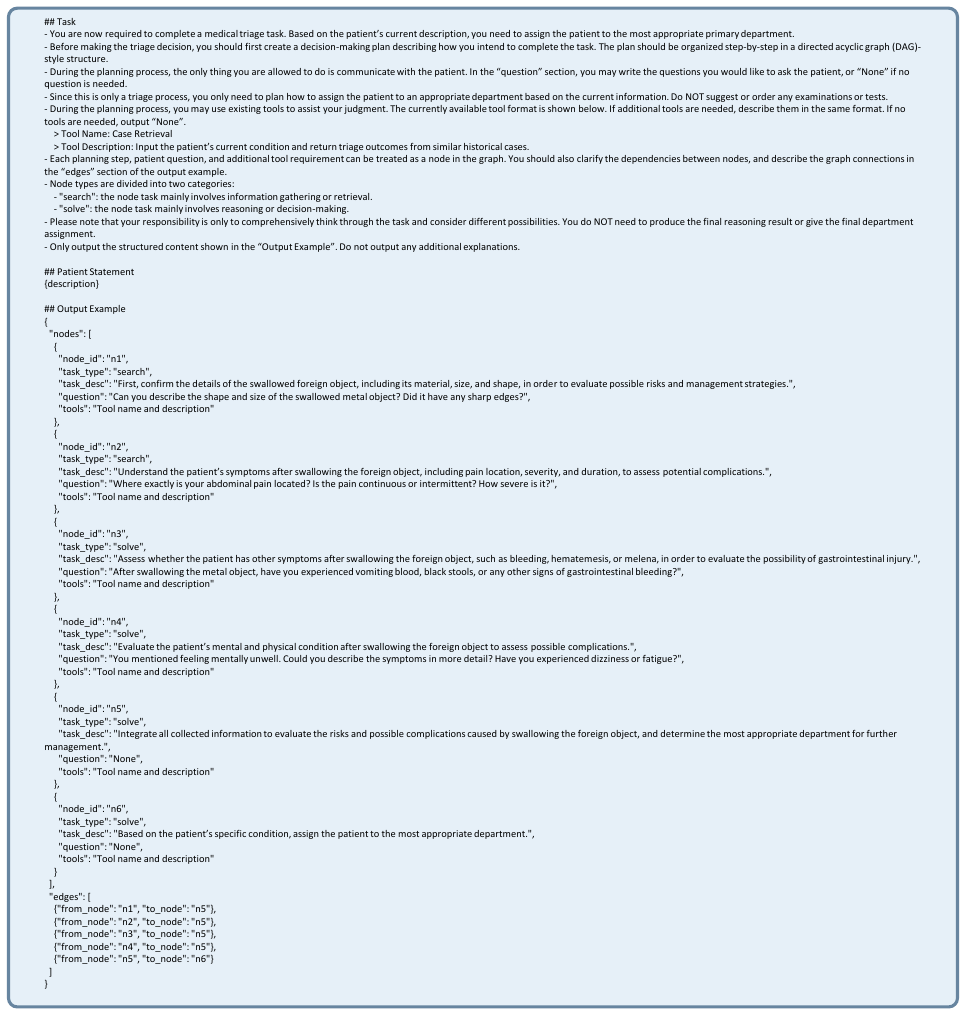}	
	\caption{The prompt of Policy Planner.}
	\label{fig_appendix:policy_planner}
\end{figure*}

\begin{figure*}[t]
	\centering
	\includegraphics[width=\textwidth]{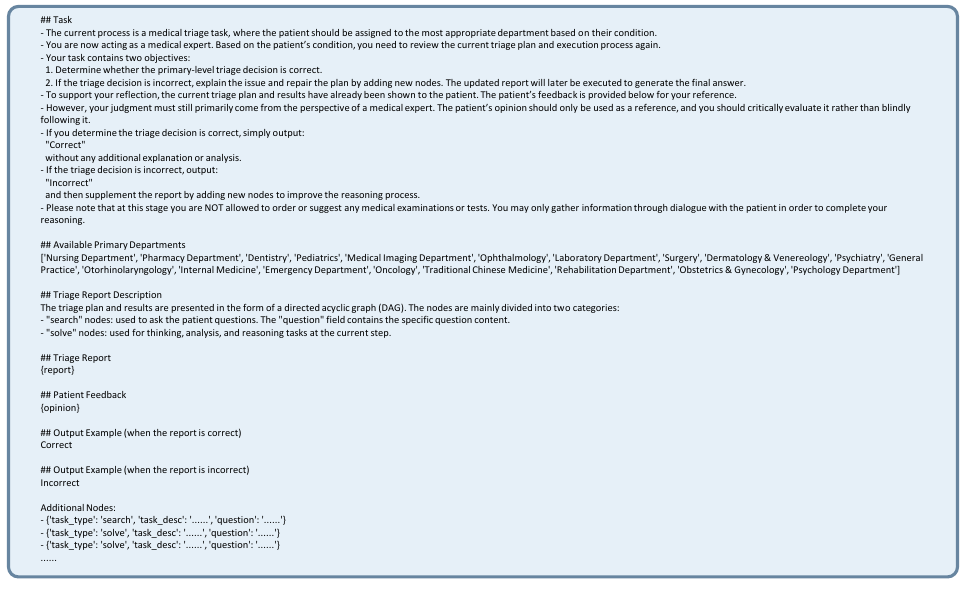}	
	\caption{The prompt of Critic Model.}
	\label{fig_appendix:critic}
\end{figure*}

\begin{figure*}[t]
	\centering
	\includegraphics[width=\textwidth]{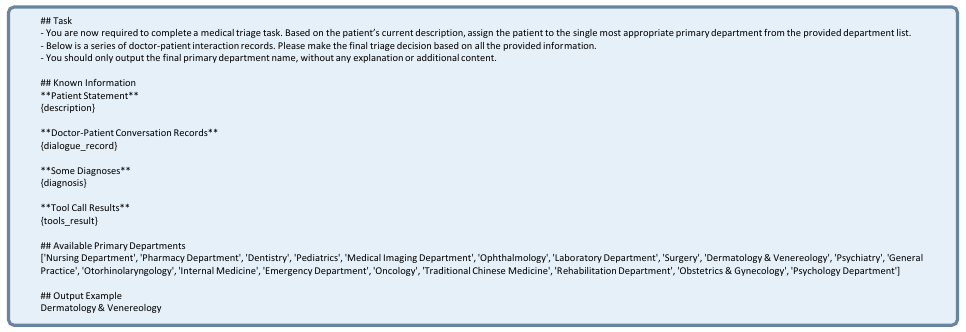}	
	\caption{The prompt of Decision Model.}
	\label{fig_appendix:decision}
\end{figure*}

In this appendix, we further present the detailed prompts used in the construction of the Doctor Agent. The Doctor Agent mainly consists of three components: the Policy Planner, the Critic Model, and the Decision Model. The following presents their specific prompts. As shown in Fig.~\ref{fig_appendix:policy_planner}, Fig.~\ref{fig_appendix:critic} and Fig.~\ref{fig_appendix:decision}.

\clearpage

\end{document}